\documentclass[runningheads]{llncs}
\usepackage[utf8]{inputenc}
\usepackage{graphicx}
\usepackage{marvosym}
\usepackage{makecell}
\usepackage[ruled,vlined,linesnumbered]{algorithm2e}
\usepackage[shortcuts]{extdash}
\usepackage{xcolor}
\usepackage[normalem]{ulem}

\begin{document}
\title{A Committee of Convolutional Neural Networks for Image Classification in the Concurrent Presence of Feature and Label Noise}
\titlerunning{Concurrent Presence of Feature and Label Noise}

\author{Stanisław Kaźmierczak$^{(\textrm{\Letter})}$\orcidID{0000-0002-8981-1592} \and \\
Jacek Ma{\'n}dziuk\orcidID{0000-0003-0947-028X}}
\authorrunning{S. Kaźmierczak and J. Ma{\'n}dziuk}
\institute{Faculty of Mathematics and Information Science, Warsaw University of Technology, Warsaw, Poland\\
\email{\{s.kazmierczak,mandziuk\}@mini.pw.edu.pl}}
\maketitle              % typeset the header of the contribution

\begin{abstract}
Image classification has become a ubiquitous task. Models trained on good quality data achieve accuracy which in some application domains is already above human-level performance. Unfortunately, real-world data are quite often degenerated by noise existing in features and/or labels. There are numerous papers that handle the problem of either feature or label noise separately. However, to the best of our knowledge, this piece of research is the first attempt to address the problem of concurrent occurrence of both types of noise. Basing on the MNIST, CIFAR-10 and CIFAR-100 datasets, we experimentally prove that the difference by which committees beat single models increases along with noise level, no matter whether it is an attribute or label disruption. Thus, it makes ensembles legitimate to be applied to noisy images with noisy labels. The aforementioned committees' advantage over single models is positively correlated with dataset difficulty level as well. We propose three committee selection algorithms that outperform a strong baseline algorithm which relies on an ensemble of individual (nonassociated) best models.

\end{abstract}

\section{Introduction}
A standard image classification task consists in assigning a correct label to an input sample picture. In the most widely-used supervised learning approach, one trains a model to recognize the correct class by providing input-output image-label pairs (training set). In many cases, the achieved accuracy is very high~\cite{chan2015pcanet,zoph2016neural}, close to or above human-level performance~\cite{ciregan2012multi,he2015delving}.
%In some tasks, AI models even outperformed humans already \cite{ciregan2012multi,he2015delving}.

The quality of real-world images is not perfect. Data may contain some noise defined as anything that blurs the relationship between the attributes of an instance and its class~\cite{hickey1996noise}. There are mainly two types of noise considered in the literature: feature (attribute) noise and class (label) noise~\cite{frenay2013classification,quinlan1986induction,zhu2004class}.

Despite the fact that machine algorithms (especially those based on deep architectures) perform on par with humans or even better on high-quality pictures, their performance on distorted images is noticeably worse~\cite{dodge2017study}. Similarly, label noise may potentially result in many negative consequences, e.g. deterioration of prediction accuracy along with an increase of model's complexity, size of a training set, or length of a training process~\cite{frenay2013classification}. Hence, it is necessary to devise methods that reduce noise or are able to perform well in its presence. The problem is furthermore important considering the fact that the acquisition of accurately labeled data is usually time-consuming, expensive and often requires a substantial engagement of human experts~\cite{breve2010semi}.

There are many papers in the literature which tackle the problem of label noise. Likewise, a lot of works have been dedicated to studying attribute noise. However, to the best of our knowledge, there are no papers that consider the problem of feature and label noise occurring simultaneously in the computer vision domain. In this paper, we present the method that successfully deals with the concurrent presence of attribute noise and class noise in image classification problem.

\subsection{The main contribution}
\label{contribution_subsection}
Encouraged by the promising results of ensemble models applied to label noise and Convolutional Neural Network (CNN) based architectures utilized to handle noisy images we examine how a committee of CNN classifiers (each trained on the whole dataset) deal with noisy images marked with noisy labels.
With regard to the common taxonomy, there are four groups of ensemble methods~\cite{kuncheva2014combining}. The first one relates to \emph{data selection mechanisms} aiming to provide different subset for every single classifier to be trained on. The second one refers to \emph{the feature level}. Methods among this group select features that each model uses. The third one, \emph{the classifier level group}, comprises algorithms that have to determine the base model, the number of classifiers, other types of classifiers, etc. The final one refers to \emph{the combination of classifiers level} where an algorithm has to decide how to combine models' individual decisions to make a final prediction. In this study, we assume having a set of well-trained CNNs which make the ultimate decision by means of soft voting (averaging) scheme~\cite{polyak1992acceleration}. We concentrate on the task of finding an optimal or near-optimal model committee that deals with concurrent presence of attribute and label noise in the image classification problem. In summary, the main contribution of this work is threefold:
\begin{itemize}
\item addressing the problem of simultaneously occurring feature and label noise which, to the best of our knowledge, is a novel unexplored setting;
\item designing three methods of building committees of classifiers which outperform a strong baseline algorithm that employs a set of individually best models;
%furthermore, their margin rises along with increasing noise.
\item proving empirically that a margin of ensembles gain over the best single model rises along with an increase of both noise types, as well as dataset difficulty, which makes the proposed approaches specifically well-suited to the case of noisy images with noisy labels.
\end{itemize}
The remainder of this paper is arranged as follows. Section~\ref{sec:literature} provides a literature review, with considerable emphasis on methods addressing label noise and distorted images. Section~\ref{sec:algorithms} introduces the proposed novel algorithms for classifiers' selection. Sections~\ref{sec:experiment} and~\ref{sec:results} describe the experimental setup and analysis of results, respectively. Finally, brief conclusions and directions for further research are presented in the last section.

\section{Related literature}
\label{sec:literature}
Many possible sources of label noise have been identified in the literature~\cite{frenay2013classification}, e.g. insufficient information provided to the expert~\cite{brodley1999identifying,frenay2013classification}, expert (human or machine) mistakes~\cite{pechenizkiy2006class,snow2008cheap}, the subjectivity of the task~\cite{hughes2004semi} or communication problems~\cite{brodley1999identifying,zhu2004class}. Generally speaking, there are three main approaches to dealing with label noise~\cite{frenay2013classification}. The first one is based on algorithms that are naturally robust to class noise. This includes ensemble methods like bagging and boosting. It has been shown in~\cite{dietterich2000experimental} that bagging performs generally better than boosting in this task. The second group of methods relies on data cleansing. In this approach, corrupted instances are identified before the training process starts off and some kind of filter (e.g. voting or partition filter~\cite{brodley1996identifying,zhu2006bridging} which is deemed easy, cheap and relatively solid) is applied to them. Ultimately, the third group consists of methods that directly model label noise during the learning phase or were specifically designed to take label noise into consideration~\cite{frenay2013classification}.

In terms of images, the key reasons behind feature noise are faults in sensor devices, analog-to-digital converter errors~\cite{shapiro2001computer} or electromechanical interferences during the image capturing process~\cite{gonzalez2008digital}. State-of-the-art approaches to deal with feature noise are founded on deep architectures. In~\cite{nazare2017deep} CNNs (LeNet-5 for MNIST and an architecture similar to the base model C of~\cite{springenberg2014striving} for CIFAR-10 and SVHN datasets) were used to handle noisy images. Application of a denoising procedure (Non-Local Means~\cite{buades2005non}) before the training phase improves classification accuracy for some types and levels of noise. In~\cite{roy2018robust} several combinations of denoising autoencoder (DAE) and CNNs were proposed, e.g. DAE-CNN, DAE-DAE-CNN, etc. The paper states that properly combined DAE and CNN achieve better results than individual models and other popular methods like Support Vector Machines~\cite{vincent2010stacked}, sparse rectifier neural network~\cite{glorot2011deep} or deep belief network~\cite{bengio2007greedy}.

\section{Proposed algorithms}
\label{sec:algorithms}
\begin{figure}[t]
\centering
\includegraphics[scale=0.30]{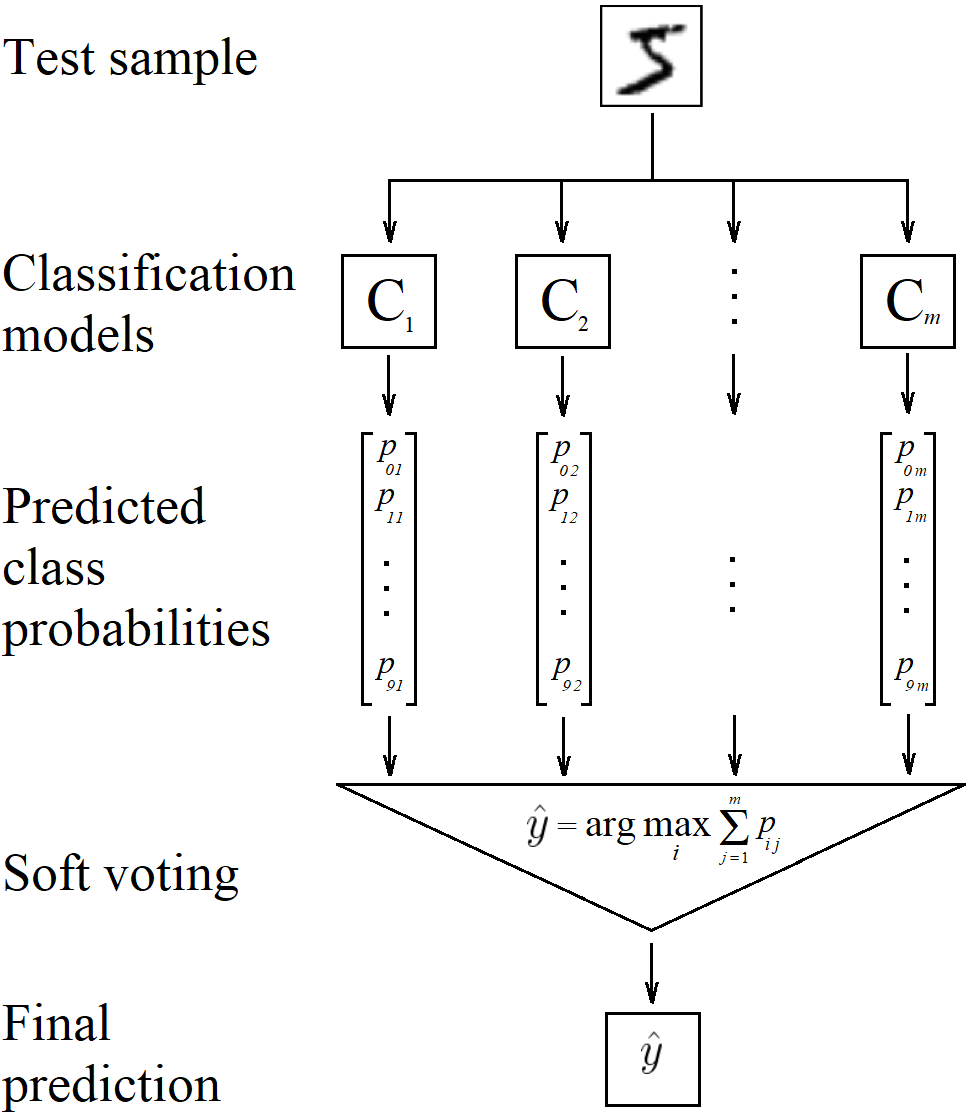}
\caption{The concept of the soft voting approach for the 10-class problem.}
\label{softvoting}
\end{figure}

As mentioned earlier, classification results are obtained via a soft voting scheme. More specifically, probabilities of particular classes from single CNNs are summed up and the class with the highest cumulative value is ultimately selected (see Fig.~\ref{softvoting}).

Many state-of-the-art results in image classification tasks are achieved by an ensemble of well-trained networks that were not selected in any way~\cite{ciregan2012multi,ioffe2015batch,krizhevsky2012imagenet}. In~\cite{simonyan2014very} the authors went further and noticed that limiting ensemble size just to two best-performing models increased accuracy in their case. We adopted that idea to the algorithm called (for the purpose of this study) \emph{top-n}, which serves as a benchmark in our experiments. First, all models are sorted in descending order according to their accuracies. Then, ensembles constituted by \emph{k} best networks where \emph{k} ranges from 1 to the number of available models are created. Finally, the committee with the best score on the validation dataset is chosen. Algorithm~\ref{top-n} summarizes the procedure.

The first algorithm proposed in this paper, called \emph{2-opt-c}, was inspired by the local search technique commonly applied to solving the Traveling Salesman Problem (TSP)~\cite{croes1958method}. The original \emph{2-opt} formulation looks for any two nodes of the current salesman's route which, if swapped, would shorten the route length. It works until no improvement is made within a certain number of sampling trials. The \emph{2-opt-c} algorithm receives an initial committee as an input
%(in its basic form it is the empty set)
and in each step modifies it by either \emph{adding} or \emph{subtracting} or \emph{exchanging} one or two elements. A single modification that maximizes accuracy on the validation set is performed. Thus, there are eight possible atomic operations listed in Algorithm~\ref{2-opt}. The procedure operates until neither of the operations improves performance. The \emph{1-opt-c} works in a very similar way but is limited to three operations that modify only one element in a committee (adding, removal and swap). The \emph{top-n-2-opt-c} and \emph{top-n-1-opt-c} operate likewise besides being initialized with the output of the \emph{top-n} procedure, not an empty committee.

\begin{algorithm}[H]
\SetKwInOut{Input}{input}\SetKwInOut{Output}{output}
\SetAlgoLined
\Input{$M_{all}$--all available models }
\Output{$C_{best}$--selected committee }
$C_{best} \leftarrow \emptyset$\;
$C_{curr} \leftarrow \emptyset$\;
$acc_{best} = 0$\;
$M_{sorted} = sort(M_{all})$\tcp*[r]{sort models descending by accuracy}
\For{$i\leftarrow 1$ \KwTo $size(M_{sorted})$}{
$C_{curr} \leftarrow C_{curr} \cup M_{sorted}[i]$\;
$acc_{curr}\leftarrow accuracy(C_{curr}) $\;
\If{$acc_{curr} > acc_{best}$}{
$acc_{best} \leftarrow acc_{curr}$\;
$C_{best} \leftarrow C_{curr}$\;
}
}
\caption{\emph{top-n}}
\label{top-n}
\end{algorithm}

\begin{algorithm}[H]
\SetKwInOut{Input}{input}\SetKwInOut{Output}{output}
\SetAlgoLined

\Input{$M_{all}$--all available models, $C_0$--initial committee }
\Output{$C_{best}$--selected committee }
$C_{best} \leftarrow C_0$\;
$acc_{best}\leftarrow accuracy(C_{best}) $\;
 \While{$acc_{best}$ rises}{
 $acc_{curr} = 0$\;
 $acc_{curr}, C_{curr} \leftarrow add(C_{best}, M_{all}, acc_{curr}) $\;
 $acc_{curr}, C_{curr} \leftarrow remove(C_{best}, acc_{curr}) $\;
 $acc_{curr}, C_{curr} \leftarrow swap(C_{best}, M_{all}, acc_{curr}) $\;
 $acc_{curr}, C_{curr} \leftarrow addTwo(C_{best}, M_{all}, acc_{curr}) $\;
 $acc_{curr}, C_{curr} \leftarrow removeTwo(C_{best}, acc_{curr}) $\;
 $acc_{curr}, C_{curr} \leftarrow addAndSwap(C_{best}, M_{all}, acc_{curr}) $\;
 $acc_{curr}, C_{curr} \leftarrow removeAndSwap(C_{best}, M_{all}, acc_{curr}) $\;
 $acc_{curr}, C_{curr} \leftarrow swapTwice(C_{best}, M_{all}, acc_{curr}) $\;
 \If{$acc_{curr} > acc_{best}$}{
 $acc_{best} \leftarrow acc_{curr}$\;
 $C_{best} \leftarrow C_{curr}$\;
}
}
\caption{\emph{2-opt-c}}
\label{2-opt}
\end{algorithm}

\begin{figure}[t]
\centering
\includegraphics[width=0.9\textwidth]{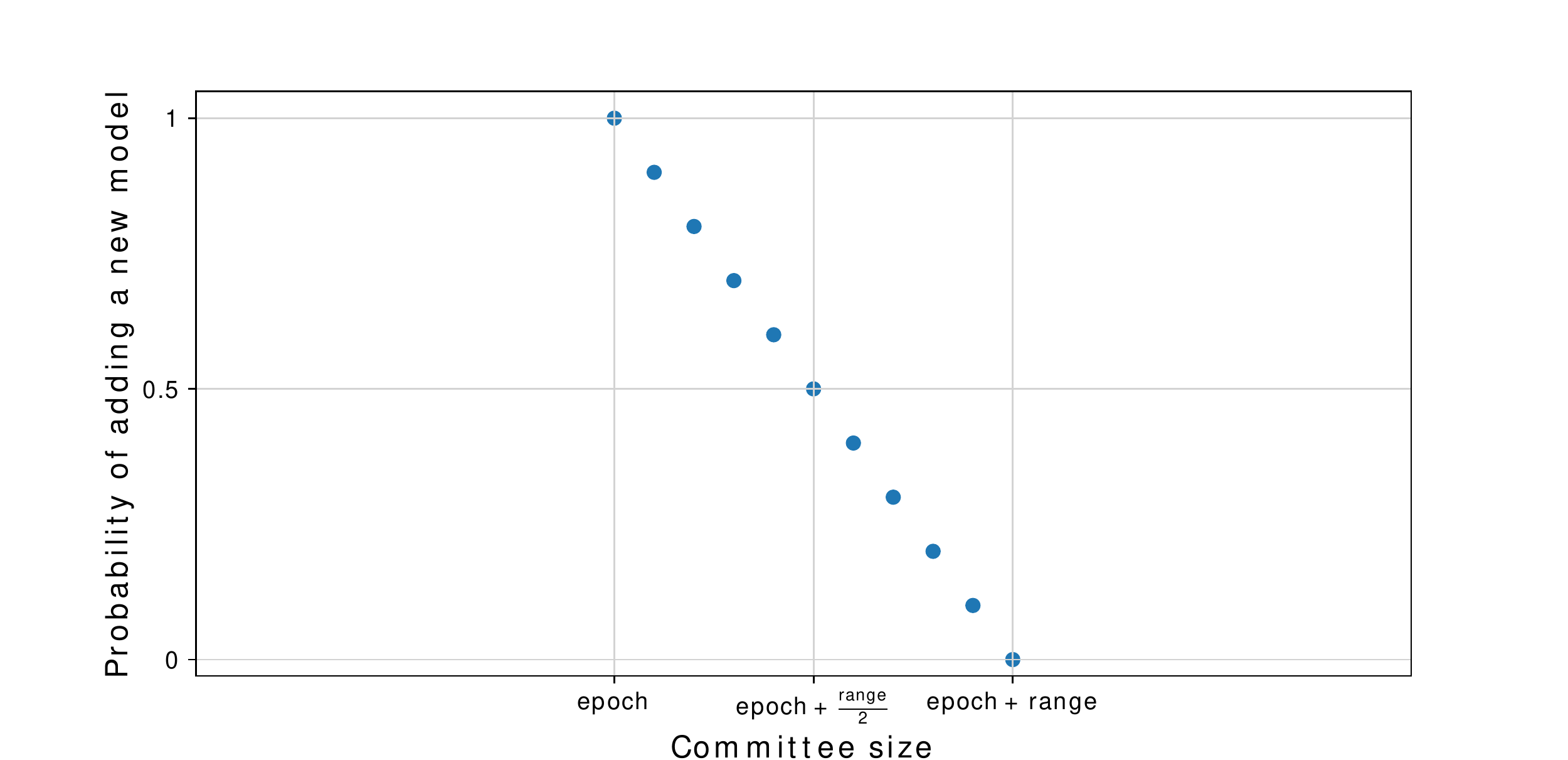}
\caption{Probability of adding a new model to the committee in the \emph{stochastic} algorithm.} \label{propability}
\end{figure}

Another algorithm, called \emph{stochastic}, relies on the fact that the performance of the entire ensemble depends on diversity among individual
%inducers
component classifiers, on the one hand, and the predictive performance of single models, on the other hand~\cite{sagi2018ensemble}. The pseudocode of the algorithm is presented in Algorithm~\ref{stochastic}. In the $i$th epoch, a committee size ranges from $i$ to $i+range$ with the expected value equal to $i+\frac{range}{2}$. This property is assured by the formula in line $7$ which increases the probability of adding a new model along with decreasing ensemble size and vice versa. Figure~\ref{propability} illustrates this relationship. In each step, one model is either added or removed. In the first scenario a model with the best individual performance is appended to the committee with probability $t_a$ (lines 11-14) or a model which minimizes the maximum correlation between any model from the committee and itself is added with probability $1-t_a$ (lines 15-18). Analogously, if the algorithm decides to decrease a committee size it removes the weakest model with probability $t_r$ (lines 22-25) or a model which minimizes the highest correlation between any two models in the committee with probability $1-t_r$ (lines 26-29). In each epoch, the algorithm performs $N_i$ iterations to explore the solution space. A correlation between two models is measured by the Pearson correlation coefficient calculated on probability vectors obtained from predictions on the validation set.

\begin{algorithm}[!tpbh]
\SetKwInOut{Input}{input}\SetKwInOut{Output}{output}
\SetAlgoLined

\Input{$M_{all}$--all available models, $N$--number of epochs, $N_i$--number of iterations within an epoch, $t_a$--probability threshold below which the strongest model is added, $t_r$--probability threshold below which the weakest model is removed, $r$--range of possible committee sizes in an epoch}
\Output{$C_{best}$--selected committee }
$C_{curr} \leftarrow \emptyset$\;
$C_{best} \leftarrow \emptyset$\;
$acc_{best}\leftarrow 0 $\;
$M_{left}\leftarrow M_{all}$\;
\For{$i\leftarrow 0$ \KwTo $N-1$}{
\For{$j\leftarrow 1$ \KwTo $N_i$}{
$p_a\leftarrow 1 - (size(C_{curr}) - i)/r $\;
$u\leftarrow$ generate from the uniform distribution $\mathcal{U}$(0, 1)\;
  \eIf{$u < p_a$}{
   $u_a\leftarrow$ generate from the uniform distribution $\mathcal{U}$(0, 1)\;
   \eIf{$size(C_{curr}) == 0$ or $u_a < t_a$}{
   $m_a\leftarrow getStrongestModel(M_{left})$\;
   $C_{curr} \leftarrow C_{curr} \cup \{m_a\}$\;
   $M_{left} \leftarrow M_{left} \setminus \{m_a\}$\;
   }{
      \tcp{select model from $M_{left}$ which minimizes maximum}
      \tcp{correlation between any model from $C_{curr}$ and itself}
        $m_a\leftarrow getMarginallyCorrelatedModel(C_{curr}, M_{left})$\;
           $C_{curr} \leftarrow C_{curr} \cup \{m_a\}$\;
   $M_{left} \leftarrow M_{left} \setminus \{m_a\}$\;
   }
   }{
   $u_r\leftarrow$ generate from the uniform distribution $\mathcal{U}$(0, 1)\;
   \eIf{$size(C_{curr}) == 1$ or $u_r < t_r$}{
   $m_r\leftarrow getWeakestModel(C_{curr})$\;
   $C_{curr} \leftarrow C_{curr} \setminus \{m_r\}$\;
   $M_{left} \leftarrow M_{left} \cup \{m_r\}$\;
   }{
      \tcp{select model from $M_{curr}$ which minimizes maximum}
      \tcp{correlation between any two models in $C_{curr}$}
        $m_r\leftarrow getMaximallyCorrelatedModel(C_{curr})$\;
   $C_{curr} \leftarrow C_{curr} \setminus \{m_r\}$\;
   $M_{left} \leftarrow M_{left} \cup \{m_r\}$\;
   }
  }
  $acc_{curr}\leftarrow accuracy(C_{curr}) $\;
\If{$acc_{curr} > acc_{best}$}{
$acc_{best} \leftarrow acc_{curr}$\;
$C_{best} \leftarrow C_{curr}$\;
}
}
}
\caption{\emph{Stochastic} algorithm}
\label{stochastic}
\end{algorithm}

\section{Experimental setup}
\label{sec:experiment}

\subsection{MNIST, CIFAR-10 and CIFAR-100 datasets}
\label{datasets_subsection}
As a benchmark, we selected three datasets with a diversified difficulty level. MNIST database contains a large set of 28x28 grayscale images of handwritten digits (10 classes) and is commonly used in machine learning experiments~\cite{lecun2010mnist}. The training set and the test set are composed of $60\,000$ and $10\,000$ images, respectively.

CIFAR-10~\cite{krizhevsky2009learning} is another popular image dataset broadly used to assess machine learning/computer vision algorithms. It contains $60\,000$ 32x32 color images in $10$ different classes. The training set includes $50\,000$ pictures, while the test set $10\,000$ ones.

The CIFAR-100 dataset is similar to CIFAR-10. It comprises $60\,000$ images with the same resolution and three color channels as well. The only difference is the number of classes--CIFAR-100 has $100$ of them, thus yielding $600$ pictures per class.

\subsection{CNN architectures}
Individual classifiers are in the form of a convolutional neural network composed of VGG blocks (i.e. a sequence of convolutional layers, followed by a max pooling layer)~\cite{simonyan2014very}, additionally enhanced by adding dropout~\cite{srivastava2014dropout} and batch normalization~\cite{ioffe2015batch}. All convolutional layers and hidden dense layer have ReLU as an activation function and their weights were initialized with \emph{He normal initializer}~\cite{he2015delving}. Softmax was applied in the output layer while the initial weights were drawn from the \emph{Glorot uniform distribution}~\cite{glorot2010understanding}. Table~\ref{tab1} summarizes the CNNs architectures. Please note that slight differences in architectures for MNIST, CIFAR-10 and CIFAR-100 are caused by distinct image sizes and numbers of classes in the datasets.
%{\color{red}Proposed architecture applied to the aforementioned benchmarks allowed us to test our algorithms on models with diversified basic performance.}
Without any attribute or label noise, a single CNN achieved approximately $99\%$, $83\%$ and $53\%$ accuracy on MNIST, CIFAR-10 and CIFAR-100, respectively.

\begin{table}[!t]
\caption{CNN architectures used for MNIST, CIFAR-10 and CIFAR-100.}\label{tab1}
\begin{tabular}[t]{|l|l|l|l|}
\hline
Layer & Type & \#maps \& neurons & kernel/pool size \\
\hline
\makecell[l]{1\\ \quad } &  \makecell[l]{convolutional\\ \quad } & \makecell[l]{32 maps of 32x32 neurons (CIFAR) \\ 32 maps of 28x28 neurons (MNIST)} & \makecell[l]{3x3\\ \quad }\\
2 & batch normalization & &\\
\makecell[l]{3\\ \quad } & \makecell[l]{convolutional\\ \quad } & \makecell[l]{32 maps of 32x32 neurons (CIFAR) \\ 32 maps of 28x28 neurons (MNIST)} & \makecell[l]{3x3\\ \quad }\\
4 & batch normalization & &\\
5 & max pooling & & 2x2\\
6 & dropout (20\%) & &\\ \hline
\makecell[l]{7\\ \quad } & \makecell[l]{convolutional\\ \quad } & \makecell[l]{64 maps of 16x16 neurons (CIFAR) \\ 64 maps of 14x14 neurons (MNIST)} & \makecell[l]{3x3\\ \quad }\\
8 & batch normalization & &\\
\makecell[l]{9\\ \quad } & \makecell[l]{convolutional\\ \quad } & \makecell[l]{64 maps of 16x16 neurons (CIFAR) \\ 64 maps of 14x14 neurons (MNIST)} & \makecell[l]{3x3\\ \quad }\\
10 & batch normalization & &\\
11 & max pooling & & 2x2\\
12 & dropout (20\%) & &\\ \hline
\makecell[l]{13\\ \quad } & \makecell[l]{convolutional\\ \quad } & \makecell[l]{128 maps of 8x8 neurons (CIFAR) \\ 128 maps of 7x7 neurons (MNIST)} & \makecell[l]{3x3\\ \quad }\\
14 & batch normalization & &\\
\makecell[l]{15\\ \quad } & \makecell[l]{convolutional\\ \quad } & \makecell[l]{128 maps of 8x8 neurons (CIFAR) \\ 128 maps of 7x7 neurons (MNIST)} & \makecell[l]{3x3\\ \quad }\\
16 & batch normalization & &\\
17 & max pooling & & 2x2\\
18 & dropout (20\%) & &\\ \hline
19 & dense & 128 neurons & \\
20 & batch normalization & &\\
21 & dropout (20\%) & &\\
\makecell[l]{22\\ \quad } & \makecell[l]{dense\\ \quad } & \makecell[l]{10 neurons (MNIST, CIFAR-10) \\ 100 neurons (CIFAR-100)} & \\
\hline
\end{tabular}
\end{table}

\subsection{Training protocol}
The following procedure was applied to all three datasets.
Partition of a dataset into training and testing subsets was predefined as described in Section~\ref{datasets_subsection}. At the very beginning, all features (RGB values) were divided by 255 to fit the $[0, 1]$ range. Images were neither preprocessed nor formatted in any other way. From the training part, we set aside $5\,000$ samples as a validation set for a single models training and another $5\,000$ samples for a committee performance comparison. From now on when referring to the training set we would mean all training samples excluding the above-mentioned $10\,000$ samples used for validation purposes.

To create noisy versions of the datasets we degraded features of the three copies of each dataset by adding Gaussian noise with standard deviation $\sigma = 0.1, 0.2, 0.3$, respectively. The above distortion was applied to the training set, two validation sets and the test set. All affected values were then clipped to $[0, 1]$ range. Next, for the original datasets and each of the three copies influenced by the Gaussian noise, another three copies were created and their training set labels were altered with probability $p = 0.1, 0.2, 0.3$, respectively. If a label was selected to be modified, a new value was chosen from the discrete uniform distribution $\mathcal{U}$\{0, 9\}. If the new label value equaled the initial value, then a new label was drawn again, until the sampled label was different from the original one. Hence, we ended up with 16 different versions of each dataset in total (no feature noise plus three degrees of feature noise multiplied by analogous four options regarding the label noise).

The second step was to train CNNs on each of the above-mentioned dataset versions. We set the maximum number of epochs to $30$ and batch size to $32$. In the case of four consecutive epochs with no improvement on the validation set, training was stopped and weights from the best epoch were restored. The Adam optimizer~\cite{kingma2014adam} was used to optimize the cross-entropy loss function: \\
$-\sum_{c=1}^{K} y_{o,c} \log(p_{o,c})$ where $K$ is the number of classes, $y_{o,c}$ -- a binary indicator whether $c$ is a correct class for observation $o$, and $p_{o,c}$ -- a predicted probability that $o$ is from class $c$. The learning rate was fixed to $0.001$.

\subsection{Algorithms parametrization}
The \emph{stochastic} algorithm was run with the following parameters: the number of available models -- $25$, the number of epochs -- $16$, the number of iterations within an epoch -- $1000$, probability threshold below which the strongest model is added -- $0.5$, probability threshold below which the weakest model is removed -- $0.5$, range of possible committee sizes in each epoch -- $10$. The above parametrization allows the algorithm to consider any possible committee size from $1$ to $25$. Other analyzed algorithms do not require setting of any steering parameters.

\section{Experimental results}
\label{sec:results}
This section presents experimental results of testing various ensemble selection algorithms. For each pair $(\sigma, p)\in\{0, 0.1, 0.2, 0.3\}\times\{0, 0.1, 0.2, 0.3\}$ $50$ CNNs were independently trained from which $25$ were drawn to create one instance of experiment. Each experiment was repeated $20$ times to obtain reliable results. In the whole study, we assume not having any knowledge regarding either the type or the level of noise the datasets are affected by.

\begin{figure}[t]
\centering
\includegraphics[width=0.9\textwidth,height=5.0cm]{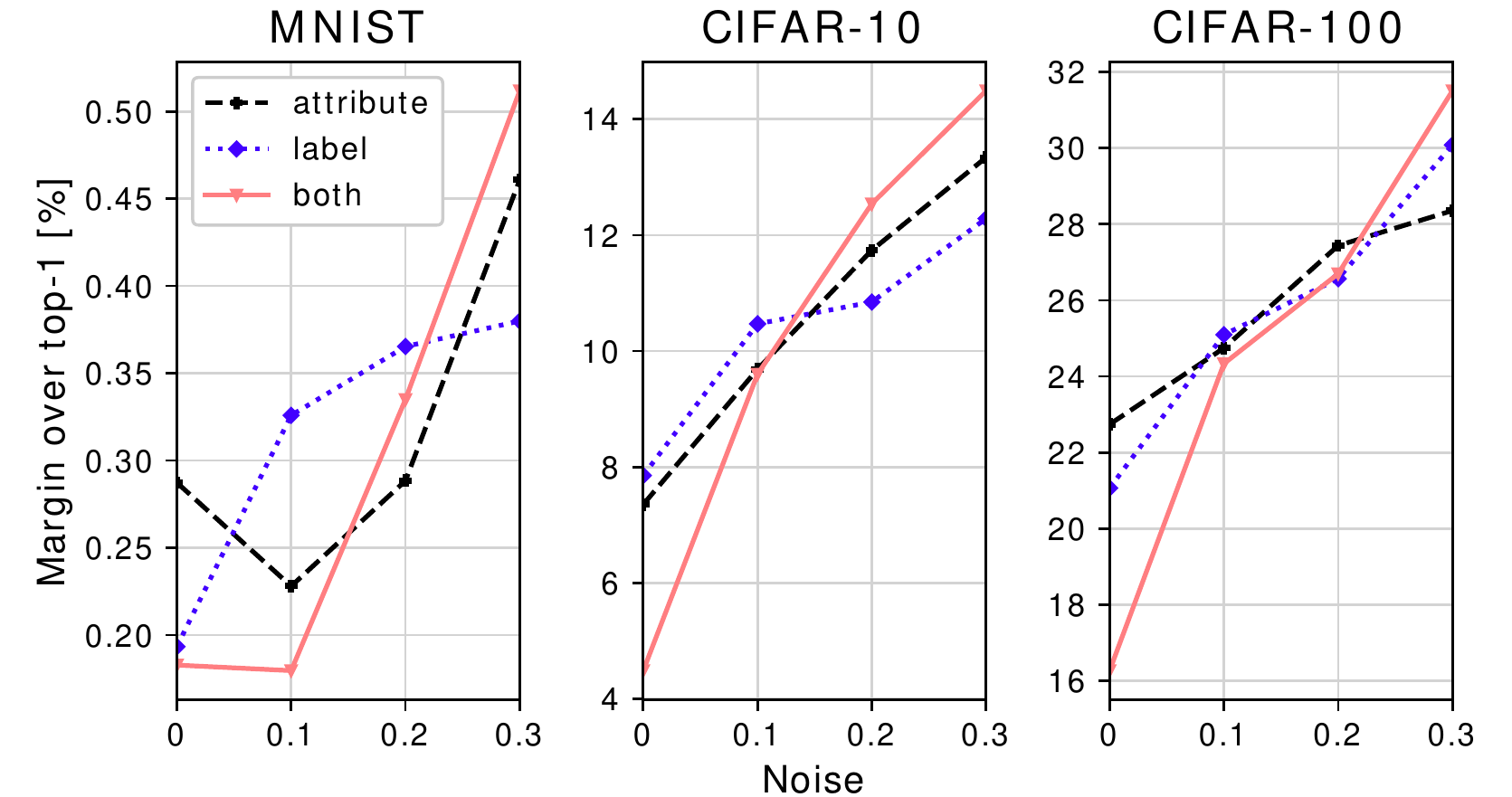}
\caption{Relative accuracy margin that committees gained over the \emph{top-1} algorithm.} \label{resultsFig1}
\end{figure}

Figure~\ref{resultsFig1} depicts the relative accuracy margin that committees gained over the \emph{top-1} algorithm which selects the best individual model from the whole library of models. Scores are averaged over \emph{top-n, 2-opt-c, 1-opt-c, top\=/n\=/2\=/opt\=/c, top\=/n\=/1\=/opt-c} and \emph{stochastic} algorithms. For example, if the best individual model achieves $80\%$ accuracy while the mean accuracy of ensembles found by analyzed algorithms equals $88\%$ then the relative margin of ensembles over \emph{top-1} is equal to $10\%$. \emph{Attribute} curves refer to computations where all scores within particular attribute noise level are averaged over label noise (four values for every attribute noise level). \emph{Label} curves are created analogously--for particular label noise all scores within specific label noise are averaged over attribute noise. \emph{Both} curves concern increasing noise level concurrently on both attributes and labels by the same amount (i.e. with $\sigma=p$). For example, $0.2$ value on the $x$\=/axis refers to $\sigma=0.2$ attribute noise and $p=0.2$ label noise.

\begin{figure}[t]
\includegraphics[width=\textwidth]{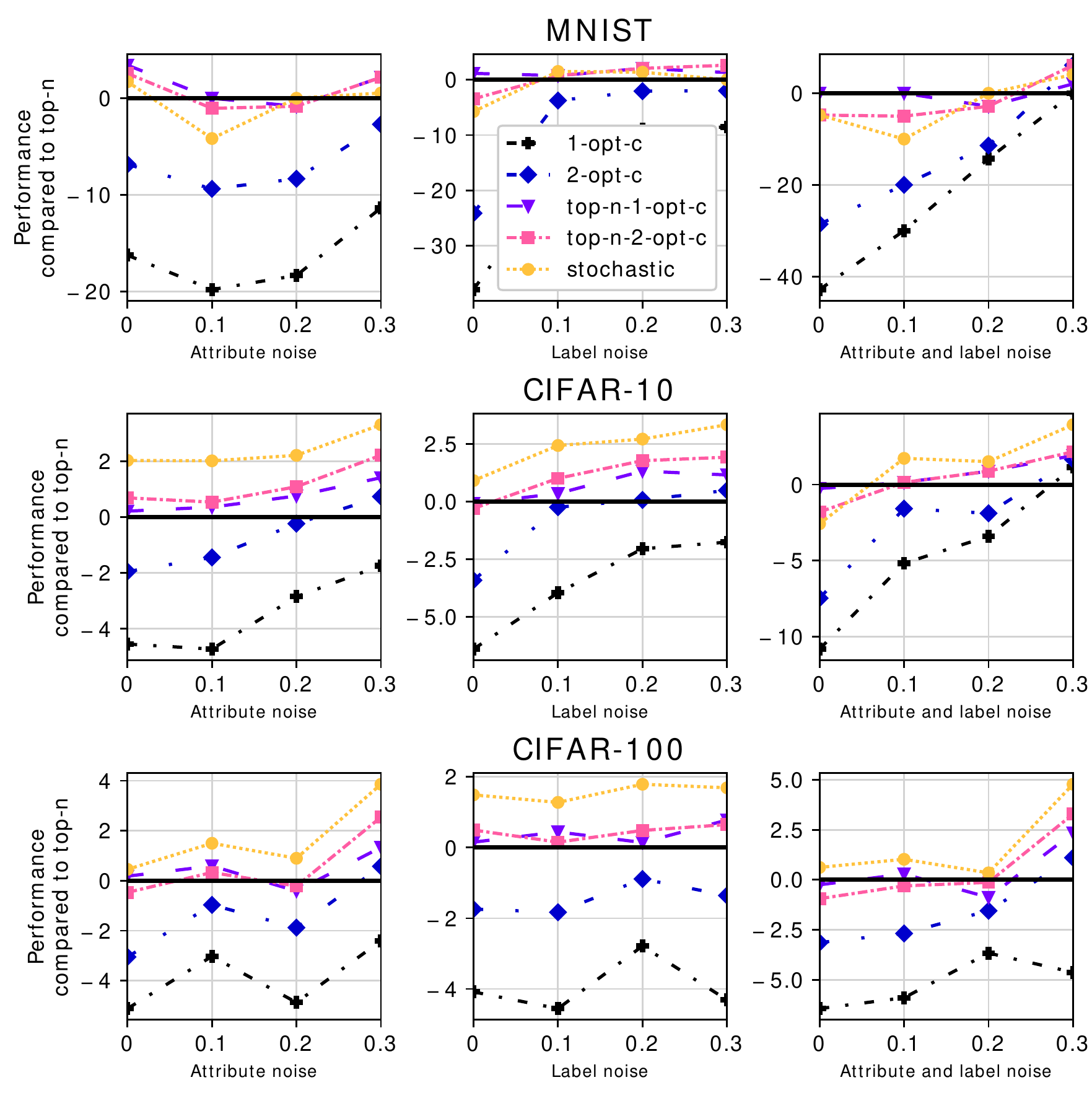}
\caption{Performance of designed algorithms contrasted with \emph{top-n} results.} \label{resultsFig2}
\end{figure}

Two main conclusions can be drawn from the plots. First, a committee margin rises along with an increase of both noise types (separately and jointly as well). Secondly, a difference increases further for more demanding datasets. In case of MNIST, the difference is less than $1\%$ while when concerning CIFAR-100 the margin amounts to more than $20\%$ for $0.1$ and $0.2$ noise level and around $30\%$ for 0.3 noise level.
Figure~\ref{resultsFig2} illustrates in a concise, aggregated way how algorithms perform in comparison with \emph{top-n} for various noise levels. Values on the $y$-axis indicate how much (percentage-wise) the margin achieved by \emph{top-n} over \emph{top\=/1} is better/worse than the margin attained by the rest of the algorithms. For example, if accuracies of \emph{top-1}, \emph{top-n} and \emph{stochastic} methods are $80\%$, $85\%$ and $85.5\%$, respectively then the value for \emph{stochastic} algorithm amounts to $10\%$ in that case since 0.5\% constitutes 10\% of 5\%. Line $y=0$ refers to \emph{top-n}. For each dataset the leftmost plot, for the given level of \emph{attribute} noise, presents scores averaged over the \emph{label} noise (four values for each level). Likewise, in the middle plot, for the given level of \emph{label} noise, the scores averaged over the four values of \emph{attribute} noise are depicted. In the third plot, the scores are not averaged since $x$-values refer to both \emph{attribute} and \emph{label} noise.
From the first row of plots, which refers to the MNIST dataset, it stems that there are huge relative differences in results achieved by the algorithms which, furthermore, vary a lot between noise levels. This phenomenon is caused by the fact that all errors for all noise levels are below $1\%$ in MNIST. Thus, even very little absolute difference between scores may be reflected in high relative value (one instance constitutes $0.01\%$ of test set size). Therefore, it is hard to draw any vital conclusions for this dataset other than a general observation that for a relatively easy dataset the results of all algorithms are close to each other.

From the plots related to CIFAR-10 and CIFAR-100, one can see that three of our algorithms noticeably surpassed the \emph{top-n} one. The \emph{stochastic} method achieved better results on all noise levels. The only yellow dot below zero refers to no noise case on either attributes and labels. Both \emph{top-n-2-opt-c} and \emph{top\=/n\=/1\=/opt\=/c} also beat \emph{top-n} in most of the cases. Another observation is that our algorithms are positively correlated with a noise level in the sense that the attained margin rises along with increasing noise.

We have also analyzed 35-sized libraries of the models. The relationships between results achieved by the algorithms remain similar to those with $25$ models, only the absolute accuracy values are slightly higher. It is not surprising since algorithms have a wider choice of models and may keep more of them in a committee. As the last remark, we noticed that \emph{2-opt-c} and \emph{1-opt-c} obtained very high accuracy on validation sets (greater than \emph{top-n-2-opt-c} and \emph{top-n-1-opt-c}, respectively) however it was not reflected on test sets. This observation suggests that one has to be careful when dealing with methods whose performance is measured solely on the validation set with neglecting models' diversity, as such committees tend to overfit. We also noticed that sizes of committees found by respective algorithms form the following order: \emph{top-n} $>$ \emph{top-n-1-opt-c} $>$ \emph{top-n-2-opt-c} $>$ \emph{stochastic} $>$ \emph{2-opt-c} $>$ \emph{1-opt-c}. Moreover, the harder dataset is the more numerous ensembles are.

\section{Conclusions and future work}
The main goal of this paper is to address the problem of concurrently occurring feature and label noise in the image classification task which, to the best of our knowledge, has not been considered in the computer vision literature.
%existing literature. 
To this end, we propose five novel ensemble selection algorithms among which four are inspired by the local optimization algorithm derived from the TSP and one employs a stochastic search. Three out of five methods outperform the strong baseline reference algorithm that applies a set of individually selected best models (\emph{top\=/n}). We have also empirically proven that a margin gained by the committees over the best single model rises along with an increase of both types of noise as well as with raising dataset difficulty, thus making proposed ensembles specifically well-suited to noisy images with noisy labels.

There are a couple of lines of inquiry worth pursuing in the future. Firstly, one may experiment with the parametrization of the \emph{stochastic} algorithm (a range of possible committee sizes in an epoch, a tradeoff between individual performance and ensemble diversity, etc.). Analysis of other correlation measures could be insightful as well. Secondly, all our algorithms operate on probability vectors, which allows us to assume that they would achieve similar results in other domains and are not limited to noisy images only. Finally, this paper addresses only one aspect of ensembling - models selection. Other areas which could be considered when forming a committee model, briefly mentioned in Section~\ref{contribution_subsection}, are also worth investigation.

\bibliographystyle{splncs04}
\bibliography{references}

\end{document}